\documentclass[9pt,a4paper,twocolumn,twoside]{rho-class/rho}
\usepackage[english]{babel}

\usepackage{amsfonts}
\usepackage{amsmath}
\usepackage{array}
\usepackage{booktabs}
\usepackage{multirow}
\usepackage{tabularx}
\usepackage{tikz}
\usepackage{wrapfig}
\usepackage{graphicx}
\usepackage{url}
\usepackage{xcolor}
\usepackage{xspace}
\usepackage[normalem]{ulem}
\usepackage{algorithm}
\usepackage{algpseudocode}
\usepackage{capt-of}
\PassOptionsToPackage{table}{xcolor}

\usepackage{amsmath,amsfonts,bm}

\def\eqref#1{equation~\ref{#1}}
\def\1{\bm{1}}

\DeclareMathAlphabet{\mathsfit}{\encodingdefault}{\sfdefault}{m}{sl}
\SetMathAlphabet{\mathsfit}{bold}{\encodingdefault}{\sfdefault}{bx}{n}

\makeatletter
\@ifundefined{citep}{\let\citep\parencite}{}
\@ifundefined{citet}{\let\citet\textcite}{}
\makeatother

\newcommand{\PaperTableStyle}{%
  \small\rmfamily
  \setlength{\tabcolsep}{5pt}%
  \renewcommand{\arraystretch}{1.10}%
}

\AtBeginEnvironment{tabular}{\PaperTableStyle}
\AtBeginEnvironment{tabular*}{\PaperTableStyle}
\AtBeginEnvironment{tabularx}{\PaperTableStyle}
\makeatletter
\g@addto@macro\normalsize{%
  \setlength{\abovedisplayskip}{12pt plus 3pt minus 2pt}%
  \setlength{\belowdisplayskip}{12pt plus 3pt minus 2pt}%
  \setlength{\abovedisplayshortskip}{9pt plus 2pt minus 2pt}%
  \setlength{\belowdisplayshortskip}{10pt plus 2pt minus 2pt}%
  \setlength{\jot}{0.45em}%
}
\makeatother

\newcolumntype{Y}{>{\raggedright\arraybackslash}X}
\definecolor{theme}{HTML}{50B4A8}

\doctype{Research Article}
\title{MemoGen: Can Past Experience Improve Future Text-to-Image Generation?}

\author[1,2,6]{Wenshuo Chen\textsuperscript{\ensuremath{\dagger}}}
\author[1]{Kuimou Yu\textsuperscript{\ensuremath{\dagger}}}
\author[1,2]{Bowen Tian\textsuperscript{\ensuremath{\dagger}}}
\author[6]{Jianfei Song\textsuperscript{\ensuremath{\dagger}}}
\author[1]{Shaofeng Liang}

\author[1]{Haozhe Jia}

\author[1,3]{Kan Cheng}
\author[1]{Haosen Li}
\author[1]{Kaishen Yuan}
\author[4,5]{Lei Wang}
\author[1,2]{Jiemin Wu}
\author[1]{Songning Lai}
\author[1,7]{Yutao Yue\textsuperscript{*}}

\affil[1]{The Hong Kong University of Science and Technology (Guangzhou)}
\affil[2]{Scholarly, Guangzhou Ziyan Technology Co., Ltd.}
\affil[3]{Shandong University}
\affil[4]{Data61/CSIRO}
\affil[5]{Griffith University}
\affil[6]{LimX Dynamics Technology Co., Ltd.}
\affil[7]{Institute of Deep Perception Technology, Jiangsu Industrial Technology Research Institute (JITRI)}

\corres{\textsuperscript{\textasteriskcentered} Corresponding author: Yutao Yue.}
\docinfo{\textsuperscript{\ensuremath{\dagger}} Equal contribution.}

\begin{abstract}
Modern text-to-image models have achieved strong visual synthesis, yet remain unreliable when prompts require implicit visual constraints, relational reasoning, or external knowledge. Existing retrieval-augmented and agentic generation methods mitigate this issue by acquiring external knowledge, references, or refined prompts for the current request, yet they typically treat each generation as an isolated episode and do not systematically preserve past successes or failures for future use. In this work, we ask whether a text-to-image system can continually improve from its own generation experience without updating the underlying generator. We propose \textbf{MemoGen}, a training-free framework that augments existing image generators with an agentic evolution layer. For each task, MemoGen explicitly infers visual requirements, retrieves external evidence and references when necessary, translates them into executable generation constraints, evaluates the generated result, and stores task understanding, reference choices, visual feedback, successful strategies, and failure lessons as reusable experience memory. Across evolution rounds, the agent retrieves relevant experience to improve similar future generations, selectively repairing previously failed cases while preserving successful ones, thereby enabling test-time self-evolution without parameter updates. Extensive experiments on knowledge-intensive and reasoning-oriented benchmarks demonstrate the effectiveness of this paradigm: after only two evolution rounds, MemoGen built upon the open-source Qwen-Image backbone surpasses strong proprietary systems such as Nano Banana Pro and GPT-Image-1 on WISE and Mind-Bench, showing that explicit experience memory can serve as a powerful continual learning signal for reliable text-to-image generation. Code will be released at \url{https://github.com/Chatonz/MemoGen}.
\end{abstract}


\begin{document}
\normalsize
\maketitle
\thispagestyle{firststyle}

\section{Introduction}

Recent advances in text-to-image generation have substantially lowered the barrier to visual content creation, enabling users to translate natural language descriptions into high-quality images\citep{rombach2022latent,
saharia2022imagen,podell2023sdxl,betker2023dalle3,chen2024sato,chen2025ant,chen2025freet2m,chen2026towards}. With the rapid development of diffusion models, unified multimodal understanding-generation models, and large-scale image generation systems, modern image generators have demonstrated increasingly strong visual synthesis capabilities. A growing body of research and system-level practice suggests that, when provided with clear, sufficiently detailed, and visually executable prompts, or appropriate reference images, current image generation models are often capable of producing outputs that closely match user expectations\citep{kou2026thinkthengeneratereasoningawaretexttoimagediffusion,feng2026gensearcher,he2026mindbrush} .In this sense, many generation failures do not necessarily stem from insufficient low-level image synthesis capability~\citep{zhang2023controlnet,mou2024t2iadapter,ye2023ipadapter,
chen2022reimagen}, but rather from the lack of accurate task understanding, external knowledge, visual references, and explicit generation constraints.

As a result, the central challenge for future image generation systems is shifting. The question is no longer only how to train more powerful generative models, but also how to automatically construct better generation conditions for already capable models. In real-world scenarios, user instructions are often brief, implicit, and underspecified. A user may describe an abstract goal without specifying the key visual cues required for faithful rendering; request a cultural, scientific, or geographical scene without providing the necessary background knowledge; or describe a task involving temporal, spatial, or physical relations that must be further reasoned about before they can be translated into visible image content. Under such conditions, an image generation system must actively acquire external information, retrieve relevant references, infer implicit visual constraints, and organize them into inputs that the underlying generator can effectively execute.

This trend has already begun to emerge in recent image generation systems. Systems such as Gen Searcher and Mind-Brush indicate\citep{feng2026gensearcher,he2026mindbrush} that augmenting image generation with web resources, visual references, and external knowledge is becoming an effective paradigm. Unlike conventional text-to-pixel models, these systems no longer treat generation as a single-step mapping process. Instead, they introduce retrieval, understanding, reasoning, and planning into the generation pipeline, making image generation closer to the workflow of human creators. When facing a complex visual subject, a human artist rarely starts from a single sentence and immediately produces a final artwork. Instead, the artist first understands the intent, searches for materials, studies references, summarizes key visual characteristics, and then creates. In an image generation system, local RAG and dynamic web search play the role of learning and information seeking: the former provides stable, efficient, and low-cost knowledge support through high-quality local databases, reducing dependence on online search, while the latter supplements open-world information when local knowledge is insufficient. Reasoning and planning then serve as summarization and conceptualization, converting external information into executable generation constraints.

However, existing retrieval-augmented or agentic image generation methods largely remain confined to single-task enhancement~\citep{li2026deltascorematters,shinn2023reflexion,park2023generative,
wang2023voyager}. Whether the information comes from local knowledge bases, reference images, or dynamic web search, the system primarily focuses on constructing better generation conditions for the current task. Once the generation is completed, successful strategies and failure cases are rarely preserved in a systematic way, nor are they reused to improve future generations. In other words, prior work has demonstrated the importance of external learning for image generation, but it has not fully answered a more fundamental question: can an image generation system continually improve from its past generation experience?

Motivated by this view, we propose \textbf{MemoGen}, a training-free continual learning framework for text-to-image generation agents~\citep{shinn2023reflexion,park2023generative,wang2023voyager}. The key insight is that a generation system can improve over time without updating the parameters of the underlying image generator. Instead, improvement can happen at the agent layer: each generation episode produces not only an image, but also task understanding, retrieved evidence, reference choices, visual feedback, and success or failure experience. These records are stored as explicit memory and later retrieved to help construct better generation conditions for similar future tasks.

At a high level, Memo turns image generation into a continual cycle of understanding, evidence acquisition, visual grounding, feedback, and experience reuse. For each prompt, the agent first makes the implicit task requirements explicit, acquires useful knowledge and references when needed, and uses them to guide an existing image generator. After generation, the system reviews whether the result satisfies the task and converts the outcome into reusable experience. Across evolution rounds, previously successful strategies can be reused, while failed cases become lessons that help avoid recurring errors. This allows the system to selectively improve difficult samples while preserving successful ones, enabling test-time self-evolution without model training.

We conduct extensive experiments to evaluate the effectiveness of MemoG en. Remarkably, after only the second evolution round, our method, built upon the open-source Qwen-Image backbone\citep{wu2025qwenimage}, already surpasses strong state-of-the-art proprietary models such as Nano Banana Pro\citep{googledeepmind2025bananapro} and GPT-Image-1\citep{openai2024gptimage}. On WISE\citep{niu2025wise}, MemoGen achieves an overall score of 0.91, outperforming Nano Banana Pro (0.87) and GPT-Image-1 (0.80), while obtaining the best results in most knowledge-intensive categories, including Cultural, Time, Chem, and Overall. Beyond WISE, we further evaluate MemoGen on Mind-Bench \citep{he2026mindbrush}, where our method reaches an overall score of 0.52, substantially exceeding Nano Banana Pro (0.41) and Mind-Brush (0.31). These results demonstrate that MemoGen can effectively improve image generation through test-time self-evolution without updating the underlying generator, enabling an open-source model to outperform leading closed-source systems in challenging knowledge-driven and reasoning-driven image generation tasks.

\noindent\textbf{Our contributions are summarized as follows:}

\begin{itemize}
    \item We propose \textbf{MemoGen}, a training-free continual learning framework for text-to-image generation agents. Instead of updating the parameters of the underlying image generator, MemoGen improves generation performance at the agent layer by storing and reusing task understanding, retrieved evidence, reference choices, visual feedback, and success or failure experiences.

    \item We formulate image generation as a \textbf{test-time self-evolution process} consisting of understanding, evidence acquisition, visual grounding, reflection, memory, and re-creation. This design enables the system to reuse successful strategies for similar future tasks while converting failed generations into lessons, thereby continuously improving difficult samples without model training.

    \item We conduct extensive experiments on \textbf{WISE} and \textbf{Mind-Bench}. After only the second evolution round, MemoGen built on the open-source Qwen-Image backbone achieves strong performance, surpassing state-of-the-art proprietary models such as Nano Banana Pro and GPT-Image-1. In particular, MemoGen achieves an overall score of 0.91 on WISE and 0.52 on Mind-Bench, demonstrating the effectiveness of training-free evolution for knowledge-driven and reasoning-driven image generation.
\end{itemize}

\begin{figure*}[t]
    \centering
    \includegraphics[width=\linewidth]{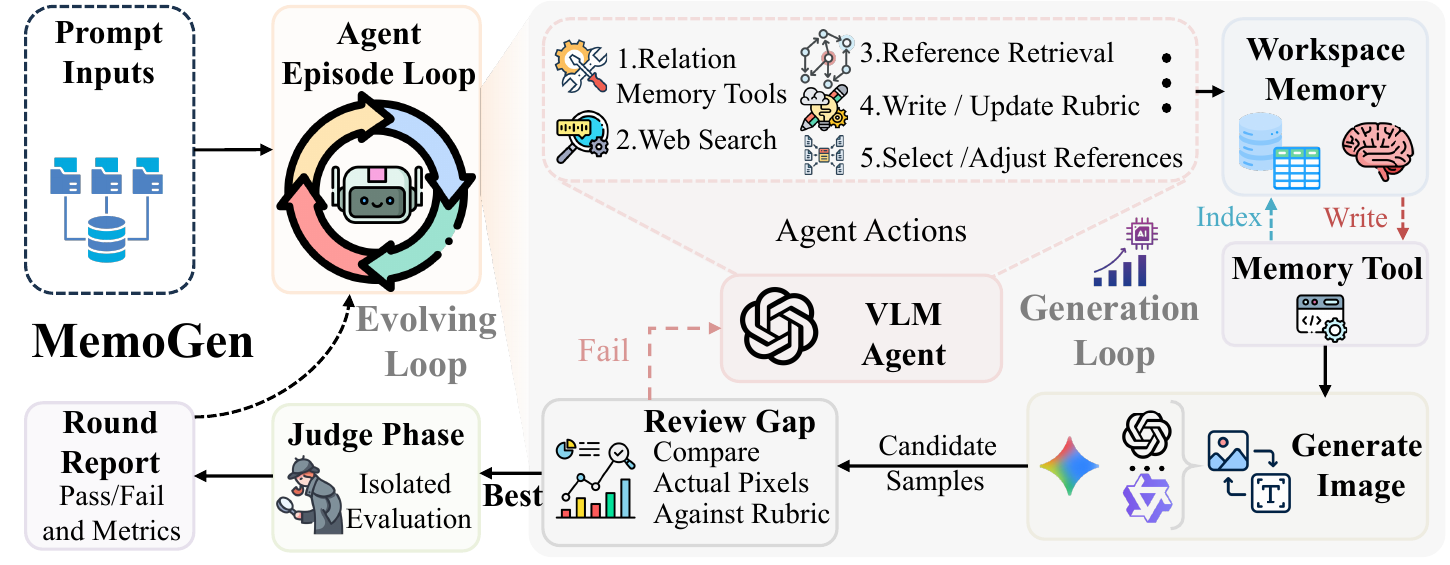}
    \caption{Overview of MemoGen. MemoGen formulates image generation as a test-time self-evolution process. This framework enables the system to continuously improve future generations by storing and reusing task understanding, retrieved evidence, reference choices, visual feedback, and success or failure experiences, all without updating the underlying model.}
    \label{fig:main}
\end{figure*}

\section{Related Work}

\paragraph{Text-to-image generation and controllable conditioning.}
Text-to-image generation has made rapid progress in recent years, driven by large-scale diffusion and autoregressive models. Latent Diffusion Models (LDMs) significantly reduce the computational cost of high-resolution synthesis by operating in a compressed latent space \citep{rombach2022latent}, while Imagen demonstrates the importance of strong language encoders for text-image alignment \citep{saharia2022imagen}. Parti further scales autoregressive generation to content-rich prompts involving complex compositions and world knowledge \citep{yu2022scalingautoregressivemodelscontentrich}, and SDXL improves latent diffusion with larger backbones and multi-aspect-ratio training \citep{podell2023sdxl}. These advances show that modern generators already possess strong visual synthesis capabilities. Alongside model scaling, ControlNet introduces spatial conditioning to pretrained diffusion models, enabling precise control with edges, depth maps, and poses \citep{zhang2023controlnet}, while T2I-Adapter learns lightweight adapters to align external control signals \citep{mou2024t2iadapter} and IP-Adapter adds image-prompt capability in a text-compatible manner \citep{ye2023ipadapter}. Retrieval-augmented methods such as kNN-Diffusion \citep{sheynin2022knndiffusion}, RA-CM3 \citep{yasunaga2023retrievalaugmentedmultimodal}, and Re-Imagen \citep{chen2022reimagen} further supplement parametric knowledge with external memory. These methods collectively suggest that, given appropriate prompts, references, or control signals\citep{zhang2023controlnet,mou2024t2iadapter,ye2023ipadapter,
chen2022reimagen}, existing generators can often produce high-quality results \citep{tian2025text2weight}. In contrast, our work focuses on how an agent can automatically construct and improve such generation conditions over time, rather than assuming they are already provided by the user or a static retrieval.

\paragraph{Agentic, feedback-driven, and memory-augmented image generation.}
Another line of work uses language or multimodal agents to plan, refine, and evaluate the generation process. T2I-Copilot introduces a training-free multi-agent system for prompt interpretation and quality evaluation \citep{chen2025t2icopilottrainingfreemultiagenttexttoimage}, while GenArtist coordinates multiple generation and editing tools with a multimodal LLM agent, decomposing complex requests into sub-problems with step-by-step verification \citep{wang2024genartistmultimodalllmagent}. Several methods introduce feedback or self-refinement into generation: VisualPrompter uses visual feedback to identify missing concepts and refine prompts accordingly \citep{wu2026visualpromptersemanticawarepromptoptimization}, RichHF shows that fine-grained human feedback improves generation quality \citep{liang2024richhumanfeedbacktexttoimage}, and Idea2Img supports multimodal iterative self-refinement via GPT-4V \citep{yang2024idea2imgiterativeselfrefinementgpt4vision}. On the search-augmented front, Gen-Searcher trains an agent via reinforcement learning to perform multi-hop search and reference selection before generation \citep{feng2026gensearcher}, while Mind-Brush formulates generation as a ``think-research-create'' cognitive workflow that retrieves multimodal evidence and reasons over implicit visual constraints \citep{he2026mindbrush}. More recently, GEMS introduces memory and reusable skills into multimodal generation \citep{he2026gems}, RationalRewards proposes a ``generate-critique-refine'' test-time loop with structured textual critiques before scoring \citep{wang2026rationalrewards}, and Maestro achieves prompt self-improvement through agent orchestration \citep{wan2025maestro}. These works improve generation through planning, feedback, search, or memory. \citep{li2026oraclenoise}In contrast, MemoGen emphasizes cross-task and cross-round experience reuse: feedback and repair knowledge from previous tasks are written into memory and later recalled to improve similar future tasks, with both successes and failures explicitly labeled for retrieval---a dimension absent from existing methods that retain only successful strategies within single-task boundaries.

\paragraph{Self-evolving agents and experience memory.}
The idea of learning from feedback without updating model parameters has been widely studied in language and embodied agents. Reflexion enables language agents to verbally reflect on task feedback and store reflective text in episodic memory, improving subsequent decision-making without gradient-based training \citep{shinn2023reflexion}. Voyager demonstrates lifelong learning in Minecraft through an ever-growing skill library, iterative prompting, and environmental feedback \citep{wang2023voyager}. While these works establish the importance of memory and reflection for language or embodied agents, T2I systems face a distinctive challenge: feedback must be converted into future visual constraints---missing objects, incorrect temporal states, ambiguous spatial relations, or insufficient domain-specific visual cues. GenEvolve, a concurrent work, first brought the self-evolution paradigm to image generation by distilling best-vs.-worst trajectory comparisons into policy parameters via visual experience distillation \citep{chen2026genevolve}. Unlike MemoGen, which keeps the generator and controller frozen, GenEvolve requires multi-stage training and its experience is implicitly dissolved in model weights, making it difficult to audit or retrieve specific past cases. EPIC addresses the feedback-to-constraint problem by compiling instructions into typed-predicate visual programs and performing targeted reconstruction on unsatisfied predicates \citep{mun2026epic}. RewardHarness validates that evolving an external tool-set while keeping the core VLM frozen can surpass fine-tuning methods requiring massive training data \citep{zhang2026rewardharness}, establishing the viability of training-free harness evolution that MemoGen extends to generative tasks. MemoGen adapts the self-evolution idea to image generation through a training-free path: the generator and controller remain fully frozen, experience is stored as polarity-labeled instance-level records in an external memory bank, and all improvement emerges from test-time accumulation and retrieval without any parameter updates.

\section{Method}

\subsection{Overview}

We propose \textbf{MemoGen}, a self-evolving image generation agent that models text-to-image generation as an experience-driven autonomous control process with tool use, visual observation, and memory writeback. Given a user instruction, an LLM/VLM-based autonomous controller manages the full generation trajectory. The controller reads the task state, visual observations, retrieval results, and past experiences; selects tool actions such as knowledge retrieval, web search, reference selection, image generation or editing, visual inspection, and memory writeback; and writes tool outputs into the next state.

Each image generation request is represented as a task
\[
x_i = (p_i, I_{\mathrm{ref},i}),
\]
where \(p_i\) is the natural-language instruction and \(I_{\mathrm{ref},i}\) denotes an optional set of user-provided reference images. Instead of treating \(x_i\) as an isolated prompt-to-image mapping, MemoGen models its execution in evolution round \(r\) as a learnable agent trajectory:
\[
\tau_i^{(r)}
=
(s_{i,0}^{(r)}, a_{i,0}^{(r)}, o_{i,0}^{(r)}, \ldots,
s_{i,T_i}^{(r)}, a_{i,T_i}^{(r)}, o_{i,T_i}^{(r)}),
\]
where \(s_{i,t}^{(r)}\) is the agent state, \(a_{i,t}^{(r)}\) is a tool action selected by the controller, and \(o_{i,t}^{(r)}\) is the corresponding tool observation. The trajectory records the original prompt, retrieved evidence, reference images, generation configuration, generated image, visual observation, internal feedback, and success or failure experience. The final image selected from this trajectory is denoted by \(\hat{I}_i^{(r)}\). In later rounds, MemoGen retrieves relevant past experiences to improve prompt construction, reference selection, tool invocation order, and failure avoidance. The system therefore accumulates test-time experience while keeping the underlying image generator frozen.

MemoGen exposes four classes of capabilities to the controller: (i) intent understanding and generation planning, which parses the instruction and identifies implicit visual constraints; (ii) knowledge and reference acquisition, which provides access to local RAG, dynamic web search, and experience memory; (iii) image generation and visual inspection, which generates candidate images and checks visible gaps; and (iv) experience memory management, which stores successful strategies, failure lessons, and reusable generation-control information. These capabilities are provided as tool interfaces to the LLM/VLM controller, and the controller determines the tool invocation order from the current state and observations. Figure~\ref{fig:main} illustrates the overall framework.

Unless otherwise specified, we use \(i\) to index tasks, \(r\) to index evolution rounds, \(k\) to index generation attempts within an episode, and \(t\) to index tool-interaction steps within an agent trajectory. We use a zero-based attempt index, where \(k=0\) denotes the initial generation attempt.

\subsection{Intent Understanding and Generation Planning}

At the beginning of each episode, MemoGen first retrieves task-relevant
experience from the current memory bank:
\[
c_{\mathcal{M},i}^{(r)}
=
\operatorname{Retrieve}(\mathcal{M}^{(r)}, p_i),
\]
where \(\mathcal{M}^{(r)}\) is the memory bank available at evolution round
\(r\), and \(c_{\mathcal{M},i}^{(r)}\) denotes the retrieved memory context for
task \(i\). Given the prompt \(p_i\) and retrieved memory context
\(c_{\mathcal{M},i}^{(r)}\), MemoGen first constructs a structured
representation of the user intent:

\[
z_{\mathrm{intent},i}^{(r)} = F_{\theta}(p_i, c_{\mathcal{M},i}^{(r)}).
\]
The intent representation includes visible objects, attributes, spatial relations, temporal states, style requirements, negative constraints, ambiguous elements, and potential knowledge dependencies.

The autonomous controller then selects or acquires a reference set:
\[
R_i^{(r,0)} = S_{\theta}(p_i, z_{\mathrm{intent},i}^{(r)}, I_{\mathrm{ref},i}, c_{\mathcal{M},i}^{(r)}),
\]
and maintains a generation plan, an initial working prompt, and an initial generation configuration:
\[
(z_{\mathrm{plan},i}^{(r)}, q_{\mathrm{work},i}^{(r,0)}, \eta_i^{(r,0)})
=
P_{\theta}(p_i, z_{\mathrm{intent},i}^{(r)}, R_i^{(r,0)}, c_{\mathcal{M},i}^{(r)}).
\]
The plan is a runtime control state that is updated as new tool observations, visual inspections, and memory retrieval results become available. During repair, the controller may update the working prompt and reference set as
\[
(q_{\mathrm{work},i}^{(r,k+1)}, R_i^{(r,k+1)}, \eta_i^{(r,k+1)})
=
U_{\theta}\!\left(
q_{\mathrm{work},i}^{(r,k)},
R_i^{(r,k)},
\eta_i^{(r,k)},
f_i^{(r,k)},
c_{\mathcal{M},i}^{(r)}
\right),
\]
where \(U_{\theta}\) denotes the controller's repair/update policy. The original prompt \(p_i\) defines the task target throughout the process. The working prompt, retrieved evidence, selected references, and memory summaries serve as auxiliary agent states that make the original instruction more visually executable.

This design targets intent execution failures in T2I tasks. Many failures arise when implicit visual constraints in underspecified user requests are left ungrounded. For example, prompts involving temporal reasoning, scientific phenomena, or cultural knowledge often require additional background knowledge before they can be translated into executable generation conditions.

\subsection{Knowledge and Reference Acquisition}

MemoGen uses three complementary tools to construct better generation
conditions.

\paragraph{Local RAG.}
Local RAG is a tool interface for image retrieval. General terminal retrieval
tools target text search; image-reference retrieval requires an additional
visual index. MemoGen therefore provides a visual-text RAG tool built on
the WIT image-text corpus~\citep{srinivasan2021wit,ning2025dctdiff}, which contains
large-scale entity-rich image-text examples from Wikipedia. Given a query, this
tool searches the local visual database and returns candidate images, related
captions, and similarity signals. It provides the controller with stable and
low-cost visual references, especially for common entities, scientific concepts,
visual states, object appearances, and recurring benchmark concepts.

\paragraph{Dynamic web search.}
When local evidence is insufficient, MemoGen invokes dynamic web or image
search to acquire open-world information. This is useful for long-tail entities,
real-time concepts, novel cultural references, and tasks with low local-index
coverage. The use of retrieved evidence and visual references is motivated by
prior retrieval-augmented generation methods, which show that external image-text
examples or multimodal neighbors can improve grounding for rare, unseen, or
out-of-distribution concepts~\citep{chen2022reimagen,sheynin2022knndiffusion}.

\paragraph{Experience memory retrieval.}
The experience memory pool stores historical generation processes as searchable
text trajectories, tool logs, feedback records, and concise experience summaries.
MemoGen provides find/grep-style terminal retrieval tools to the VLM
controller, allowing the controller to directly search the experience memory
pool and locate relevant failure modes, successful strategies, reference choices,
and repair records. This retrieval mode follows the file-level search pattern
commonly used by tool-using agents: the controller can iterate on queries, read
local snippets, and select interpretable evidence before generation. This design
is also inspired by memory-based agent learning, where previous feedback or
experience is stored externally and later reused without parameter
updates~\citep{shinn2023reflexion,wang2023voyager}.

These information sources are exposed as callable tools to the LLM/VLM
controller. Textual experience is retrieved through find/grep-style tools, while
visual references are retrieved through local RAG or web image search. Based on
the task description, current retrieval results, visual observations, and past
memory, the controller decides when additional knowledge is needed, when web
search should be invoked, when reference images should be supplemented, and when
retrieval should stop before generation. The resulting context is summarized
into \(c_{\mathcal{M},i}^{(r)}\) and the selected reference set
\(R_i^{(r,k)}\), which are then used by the planner and generator. This
tool-selection mechanism leverages the visual understanding and cross-modal
reasoning capability of current VLMs, enabling flexible context completion and
evidence integration in open-world generation tasks.

\subsection{Image Generation and Visual Inspection}

Given the current working prompt, reference set, and generation configuration, MemoGen calls an image generation or editing backend:
\[
I_i^{(r,k)} = G_{\phi}\!\left(q_{\mathrm{work},i}^{(r,k)}, R_i^{(r,k)}, \eta_i^{(r,k)}\right),
\]
where \(I_i^{(r,k)}\) is the \(k\)-th generated attempt for task \(i\) in round \(r\). The generated image enters a visual inspection stage:
\[
O_i^{(r,k)} = H_{\psi}\!\left(I_i^{(r,k)}, p_i, z_{\mathrm{intent},i}^{(r)}, q_{\mathrm{work},i}^{(r,k)}\right),
\]
where \(O_i^{(r,k)}\) contains visible objects, attributes, spatial relations, missing elements, uncertain regions, and obvious artifacts. When necessary, the system may create resized views or cropped regions to inspect fine-grained details. These visual inspection operations support runtime attention allocation and help the agent estimate task satisfaction and select repair actions.

The gap review module produces an internal feedback record:
\[
f_i^{(r,k)} = V_{\theta}\!\left(
 p_i,
 z_{\mathrm{intent},i}^{(r)},
 z_{\mathrm{plan},i}^{(r)},
 q_{\mathrm{work},i}^{(r,k)},
 I_i^{(r,k)},
 O_i^{(r,k)}
\right).
\]
Each feedback record summarizes the satisfied requirements, violated constraints, incorrect or extra visual elements, reference-use issues, and a repair recommendation. We denote its discrete outcome as
\[
y_i^{\mathrm{int},(r,k)} \in \{\mathrm{sat}, \mathrm{unsat}, \mathrm{uncertain}\},
\]
where \(\mathrm{sat}\) indicates that the generated image satisfies the current rubric, \(\mathrm{unsat}\) indicates that one or more task constraints are violated, and \(\mathrm{uncertain}\) indicates that additional visual inspection is required. This feedback provides the control signal for local repair, additional inspection, and memory writeback.
\begin{algorithm}[t]
\scriptsize
\caption{\textsc{MemoGen}: Training-free Self-Evolution}
\label{alg:genevolving}
\begin{algorithmic}[1]
\Require Dataset \(\mathcal{D}=\{(p_i,I_{\mathrm{ref},i})\}_{i=1}^{N}\); memory \(\mathcal{M}^{(0)}\); frozen generator \(G_{\phi}\); modules \(F_{\theta},S_{\theta},P_{\theta},H_{\psi},V_{\theta},U_{\theta}\); independent memory judge \(J_{\mathrm{mem}}\); rounds \(N_{\mathrm{round}}\); max attempts \(K_{\max}\).
\Ensure Final images \(\{\hat{I}_i^{(N_{\mathrm{round}}-1)}\}_{i=1}^{N}\) and memory \(\mathcal{M}^{(N_{\mathrm{round}})}\).

\For{\(r=0,\ldots,N_{\mathrm{round}}-1\)}
    \State \(\mathcal{C}^{(r)} \gets \emptyset\)
    \For{each \(x_i=(p_i,I_{\mathrm{ref},i}) \in \mathcal{D}\)}
        \State \(c_{\mathcal{M}} \gets \operatorname{Retrieve}(\mathcal{M}^{(r)},p_i)\)
        \State \(z_{\mathrm{intent}} \gets F_{\theta}(p_i,c_{\mathcal{M}})\)
        \State \(R^{0} \gets S_{\theta}(p_i,z_{\mathrm{intent}},I_{\mathrm{ref},i},c_{\mathcal{M}})\)
        \State \((z_{\mathrm{plan}},q^{0},\eta^{0}) \gets P_{\theta}(p_i,z_{\mathrm{intent}},R^{0},c_{\mathcal{M}})\)
        \State \(k \gets 0\), \(\mathrm{finished} \gets \mathrm{false}\)

        \While{\(k<K_{\max}\) \textbf{and} \(\neg \mathrm{finished}\)}
            \State \(I^{k} \gets G_{\phi}(q^{k},R^{k},\eta^{k})\)
            \State \(O^{k} \gets H_{\psi}(I^{k},p_i,z_{\mathrm{intent}},q^{k})\)
            \State \(f^{k} \gets V_{\theta}(p_i,z_{\mathrm{intent}},z_{\mathrm{plan}},q^{k},I^{k},O^{k})\)

            \If{finish \textbf{or} \(k=K_{\max}-1\)}
                \State \(\mathrm{finished} \gets \mathrm{true}\)
            \Else
                \State \((q^{k+1},R^{k+1},\eta^{k+1}) \gets U_{\theta}(q^{k},R^{k},\eta^{k},f^{k},c_{\mathcal{M}})\)
                \State \(k \gets k+1\)
            \EndIf
        \EndWhile

        \State \(K_i^{(r)} \gets k+1\)
        \State Select \(k_i^{\star} \in \{0,\ldots,K_i^{(r)}-1\}\)
        \State \(\hat{I}_i^{(r)} \gets I^{k_i^{\star}}\)
        \State \(c_i^{\mathrm{int},(r)} \gets W_{\mathrm{int}}(\tau_i^{(r)},\{f^{k}\}_{k=0}^{K_i^{(r)}-1})\)
        \State \((y_i^{\mathrm{mem},(r)},d_i^{\mathrm{mem},(r)}) \gets J_{\mathrm{mem}}(p_i,\hat{I}_i^{(r)},I_{\mathrm{ref},i},R^{k_i^\star},c_{\mathcal{M}},z_{\mathrm{intent}},z_{\mathrm{plan}})\)
        \State \(c_i^{\mathrm{judge},(r)} \gets W_{\mathrm{judge}}(y_i^{\mathrm{mem},(r)},d_i^{\mathrm{mem},(r)},\tau_i^{(r)},c_i^{\mathrm{int},(r)})\)
        \State \(\mathcal{C}^{(r)} \gets \mathcal{C}^{(r)} \cup \{c_i^{\mathrm{int},(r)},c_i^{\mathrm{judge},(r)}\}\)
    \EndFor
    \State \(\mathcal{M}^{(r+1)} \gets \mathcal{M}^{(r)} \cup \mathcal{C}^{(r)}\)
\EndFor

\State \Return \(\{\hat{I}_i^{(N_{\mathrm{round}}-1)}\}_{i=1}^{N}\), \(\mathcal{M}^{(N_{\mathrm{round}})}\)
\end{algorithmic}
\end{algorithm}
\subsection{Experience Memory}

The key component of MemoGen is experience memory. After valid internal feedback is produced, the system writes the current episode into memory:
\[
c_i^{\mathrm{int},(r)}
=
W_{\mathrm{int}}\!\left(\tau_i^{(r)}, \{f_i^{(r,k)}\}_{k=0}^{K_i^{(r)}-1}\right).
\]
Each memory case stores the original prompt \(p_i\), working prompts \(\{q_{\mathrm{work},i}^{(r,k)}\}\), retrieved memory context \(c_{\mathcal{M},i}^{(r)}\), selected references \(\{R_i^{(r,k)}\}\), generation configurations \(\{\eta_i^{(r,k)}\}\), generated image path, visual observations \(\{O_i^{(r,k)}\}\), internal feedback \(\{f_i^{(r,k)}\}\), and a concise experience summary. Rather than storing raw free-form reflections as unstructured prompt text, MemoGen converts feedback into reusable experience records that can later be retrieved as relation-level guidance for similar tasks.

We denote the attempt-level records in round \(r\) as
\[
\mathcal{Q}_i^{(r)}
=
\{q_{\mathrm{work},i}^{(r,k)}\}_{k=0}^{K_i^{(r)}-1},
\quad
\mathcal{R}_i^{(r)}
=
\{R_i^{(r,k)}\}_{k=0}^{K_i^{(r)}-1},
\]
\[
\mathcal{E}_i^{(r)}
=
\{\eta_i^{(r,k)}\}_{k=0}^{K_i^{(r)}-1},
\quad
\mathcal{O}_i^{(r)}
=
\{O_i^{(r,k)}\}_{k=0}^{K_i^{(r)}-1},
\quad
\mathcal{F}_i^{(r)}
=
\{f_i^{(r,k)}\}_{k=0}^{K_i^{(r)}-1}.
\]
A memory record can then be abstractly represented as
\[
\begin{aligned}
c_i^{\mathrm{int},(r)}
=
\big\{
& p_i,\,
c_{\mathcal{M},i}^{(r)},\,
\mathcal{Q}_i^{(r)},\,
\mathcal{R}_i^{(r)},\,
\mathcal{E}_i^{(r)},\,
\hat{I}_i^{(r)}, \\
& \mathcal{O}_i^{(r)},\,
\mathcal{F}_i^{(r)},\,
\ell_i^{(r)},\,
\rho_i^{(r)},\,
\pi_i^{(r)}
\big\}.
\end{aligned}
\]
where \(\ell_i^{(r)}\) is the textual lesson, \(\rho_i^{(r)} \in \{\mathrm{positive},\mathrm{negative}\}\) is the memory polarity, and \(\pi_i^{(r)} \in \{\mathrm{reuse\_as\_strategy},\mathrm{use\_as\_warning}\}\) is the retrieval policy. The polarity is assigned from the internal feedback \(f_i^{(r,k)}\) and the independent memory-judge signal \(y_i^{\mathrm{mem},(r)}\), not from official benchmark labels or hidden checklist annotations.

Each experience record is associated with a polarity that determines how it is used in future episodes. \textbf{Positive records} capture visual constraints, reference choices, prompt formulations, and retrieved evidence that led to a satisfactory generation under the memory judge. They can be reused as candidate strategies for constructing future generation conditions. \textbf{Negative records} capture violated constraints and recurring failure patterns, such as missing visual cues, incorrect temporal states, ambiguous object relations, reference misuse, insufficient domain-specific depiction, or a critical failure detected by the memory judge. They are retrieved as warnings or repair constraints, preventing the controller from treating failed cases as positive prompt templates.

\begin{table*}[t]
\centering
\small
\setlength{\tabcolsep}{5.2pt}
\renewcommand{\arraystretch}{1.06}
\caption{Comparison with existing image generation models on WISE. The best performing model in each column is highlighted in \textbf{bold}.}
\label{tab:wise_results}
\resizebox{\textwidth}{!}{
\begin{tabular}{l|ccccccc}
\toprule
\multirow{2}{*}{\textbf{Model Name}}
& \multicolumn{7}{c}{\textbf{WISE}} \\
\cmidrule(lr){2-8}
& Cultural & Time & Space & Bio & Phys & Chem & Overall \\
\midrule

GPT-Image-1~\citep{openai2024gptimage}
& 0.81 & 0.71 & 0.89 & 0.83 & 0.79 & 0.74 & 0.80 \\

Nano Banana~\citep{googledeepmind2025banana}
& 0.89 & 0.87 & \textbf{0.95} & \textbf{0.89} & \textbf{0.89} & 0.79 & 0.89 \\

Nano Banana Pro~\citep{googledeepmind2025bananapro}
& 0.89 & 0.80 & 0.89 & 0.88 & 0.86 & \textbf{0.85} & 0.87 \\

\midrule

FLUX.1-dev~\citep{labs2024flux}
& 0.48 & 0.58 & 0.62 & 0.42 & 0.51 & 0.35 & 0.50 \\

SD-XL-base~\citep{podell2023sdxl}
& 0.43 & 0.48 & 0.47 & 0.44 & 0.45 & 0.27 & 0.43 \\

SD-3.5-medium~\citep{stabilityai2024sd35medium}
& 0.43 & 0.50 & 0.52 & 0.41 & 0.53 & 0.33 & 0.45 \\

SD-3.5-large~\citep{stabilityai2024sd35large}
& 0.44 & 0.50 & 0.58 & 0.44 & 0.52 & 0.31 & 0.46 \\

BAGEL (w/ CoT)~\citep{deng2025bagel}
& 0.76 & 0.69 & 0.75 & 0.65 & 0.75 & 0.58 & 0.70 \\

BAGEL~\citep{deng2025bagel}
& 0.44 & 0.55 & 0.68 & 0.44 & 0.60 & 0.39 & 0.52 \\

Qwen-Image~\citep{wu2025qwenimage}
& 0.62 & 0.63 & 0.77 & 0.57 & 0.75 & 0.40 & 0.62 \\

GenAgent~\citep{jiang2026genagent}
& 0.78 & 0.67 & 0.78 & 0.72 & 0.77 & 0.55 & 0.72 \\

Gen-Searcher-8B + Qwen-Image~\citep{feng2026gensearcher}
& 0.80 & 0.71 & 0.82 & 0.76 & 0.74 & 0.75 & 0.77 \\

Mind-Brush~\citep{he2026mindbrush}
& 0.83 & 0.69 & 0.84 & 0.71 & 0.85 & 0.68 & 0.78 \\

\rowcolor{green!8}
\textbf{MemoGen (Ours)}
& \textbf{0.96} & \textbf{0.90} & 0.94 & 0.82 & 0.83 & \textbf{0.85} & \textbf{0.91} \\

\bottomrule
\end{tabular}
}
\end{table*}

\subsection{Two-Stage Feedback Memory}

MemoGen constructs experience memory from two feedback stages. The first stage is internal visual feedback produced within the generation episode. After a generated image is returned to the controller as visual context, the controller compares the actual image against the original instruction, the inferred intent, the current plan, and the current working prompt. This produces the attempt-level feedback \(f_i^{(r,k)}\), which records visible constraint satisfaction, missing elements, reference-use issues, artifacts, and possible repair directions. Since this feedback is derived only from controller-visible information, it can be used for within-episode control, such as additional inspection, prompt repair, or memory writeback.

The second stage is an independent memory-judge feedback signal produced after an episode is completed. Instead of using official benchmark correctness labels, official judge rationales, hidden annotations, or checklist-level ground truth, 
we introduce an independent episode-level multimodal memory evaluator
\(J_{\mathrm{mem}}\) dedicated to memory writing.
 It evaluates only the original instruction, the final generated image, optional user-provided references, selected reference images, retrieved evidence, intent analysis, and generation plan:
\[
\left(y_i^{\mathrm{mem},(r)}, d_i^{\mathrm{mem},(r)}\right)
=
J_{\mathrm{mem}}\!\left(
p_i,
\hat{I}_i^{(r)},
I_{\mathrm{ref},i},
R_i^{(r,k_i^\star)},
c_{\mathcal{M},i}^{(r)},
z_{\mathrm{intent},i}^{(r)},
z_{\mathrm{plan},i}^{(r)}
\right).
\]
Here \(y_i^{\mathrm{mem},(r)} \in \{0,1\}\) indicates whether the final image is accepted by the memory judge, and \(d_i^{\mathrm{mem},(r)}\) is a structured diagnosis used for writing interpretable memory.

The memory evaluator follows five task-visible criteria: original-instruction faithfulness, intent-requirement satisfaction, reasoning or knowledge correctness, visual decidability and reference fidelity, and the presence of any critical failure. A generation is marked positive only when the judge can verify from the visible image and allowed context that the core task requirements are satisfied. If the image violates the original instruction, misses a key inferred intent, contradicts retrieved knowledge, fails to preserve required references, or contains a critical visible failure, the memory signal is negative. This design makes the memory signal independent from benchmark-specific annotations while still providing useful success/failure supervision for self-evolution.

Given \(y_i^{\mathrm{mem},(r)}\), \(d_i^{\mathrm{mem},(r)}\), the agent-visible trajectory, and the internal feedback memory, MemoGen writes a judge-conditioned memory record:
\[
c_i^{\mathrm{judge},(r)}
=
W_{\mathrm{judge}}\!\left(
y_i^{\mathrm{mem},(r)},
d_i^{\mathrm{mem},(r)},
\tau_i^{(r)},
c_i^{\mathrm{int},(r)}
\right).
\]
The resulting record preserves or revises the control experience accumulated in the episode, including prompt strategies, reference choices, tool-use patterns, failure diagnoses, and failure-avoidance rules.

This interface can be viewed as a lightweight feedback channel for image generation agents. Unlike standard RLHF\citep{ouyang2022rlhf} or preference optimization, MemoGen does not update the parameters of the image generator or controller. Instead, the memory-judge outcome is converted into explicit memory that influences future test-time behavior through retrieval. In interactive deployment, \(J_{\mathrm{mem}}\) can be replaced or supplemented by user acceptance, rejection, or preference feedback, but in benchmark evaluation the official benchmark evaluator is used only for offline reporting and is never exposed to the self-evolution memory.

\subsection{Self-Evolving Generation Loop}

MemoGen operates over multiple evolution rounds. In the initial round, the agent performs cold-start generation without task-specific experience memory. Each episode produces an image, internal visual feedback, and an internal experience record. After the episode is completed, the independent memory judge \(J_{\mathrm{mem}}\) produces a memory-writing signal and structured diagnosis, which are converted into judge-conditioned memory.

In later rounds, the controller retrieves relevant experience records before constructing the generation plan. Positive records provide candidate strategies, such as effective visual constraints, reference choices, prompt formulations, and useful evidence. Negative records provide warnings or repair constraints that help avoid recurring failure modes. The agent then executes a new generation episode under the same controlled tool interface, and the newly produced internal feedback and memory-judge feedback are written back into the experience memory.

The update across a complete evolution round is
\[
\mathcal{C}^{(r)}
=
\bigcup_{i=1}^{N}
\left\{
c_i^{\mathrm{int},(r)},
c_i^{\mathrm{judge},(r)}
\right\},
\qquad
\mathcal{M}^{(r+1)}
=
\mathcal{M}^{(r)} \cup \mathcal{C}^{(r)}.
\]
This loop enables training-free self-evolution: the image generator remains fixed, while the agent changes its future behavior through memory retrieval and memory writeback. Learning therefore occurs through updates to the external experience state rather than through model-parameter updates.

In our benchmark implementation, each evolution round reruns the same task set under the autonomous control framework. The original prompt is always kept as the task target across rounds; retrieved evidence, enhanced prompts, plans, and memory records serve only as auxiliary agent states. Reusing the same task set lets us measure whether accumulated experience improves future generations under a controlled and comparable setting. Crucially, the cross-round signal is produced by internal visual feedback and the independent memory judge, not by official benchmark correctness labels, hidden annotations, official rationales, or checklist-level ground truth.

Algorithm~\ref{alg:genevolving} summarizes the full procedure. For compactness, within each task and round, Algorithm~\ref{alg:genevolving} abbreviates \(q_{\mathrm{work},i}^{(r,k)}, R_i^{(r,k)}, \eta_i^{(r,k)}, I_i^{(r,k)}, O_i^{(r,k)}, f_i^{(r,k)}\) as \(q^k, R^k, \eta^k, I^k, O^k, f^k\), respectively.

\subsection{Scalable Benchmark Instantiation}

For large-scale evaluation, we instantiate MemoGen as a reproducible autonomous control framework. The framework provides a unified tool interface to the LLM/VLM controller, including prompt reasoning, memory retrieval, local RAG, web search, reference-image acquisition, image generation or editing, visual inspection, memory-judge feedback, and memory writeback. For each sample, the controller autonomously determines the order of tool calls from the current state and sends the generated image to the independent memory judge \(J_{\mathrm{mem}}\). The memory judge outputs only a binary memory signal and a structured diagnosis based on allowed inputs.

We do not use official benchmark correctness labels, official judge rationales, hidden annotations, or checklist-level ground truth as self-evolution signals. The memory signal is produced by the independent calibrated multimodal judge, which evaluates only the original instruction, generated image, optional references, selected references, retrieved evidence, intent analysis, and generation plan. Official benchmark evaluators are invoked only after each round for offline reporting of final metrics, and their labels, rationales, checklist decisions, or category-wise correctness signals are never written into the harness memory or shown to the controller.

This benchmark instantiation preserves the core self-evolving mechanism while placing unstable text-to-image and image-to-image backends under the control of an explicit harness memory. Each round reruns the autonomous control process for all samples and writes both internal feedback memory and independent judge-conditioned reflection memory, allowing the system to continually update prompt strategies, reference selection, visual inspection experience, and failure-avoidance rules over the full distribution. Across benchmarks, the original prompt or instruction is always kept as the task target; enhanced prompts, plans, retrieved knowledge, and memory summaries serve only as auxiliary agent states.

\begin{table*}[t]
\centering
\small
\setlength{\tabcolsep}{4.2pt}
\renewcommand{\arraystretch}{1.08}
\caption{Quantitative comparison of different models on Mind-Bench. The table is divided into proprietary (top) and open-source (bottom) models. The best performing model is highlighted in \textbf{bold}. The symbol ``--'' indicates that the model is not applicable to certain tasks.}
\label{tab:mindbench_results}
\resizebox{\textwidth}{!}{
\begin{tabular}{l|ccccc|ccccc|c}
\toprule
\multirow{2}{*}{\textbf{Model Name}}
& \multicolumn{5}{c|}{\textbf{Knowledge-Driven}}
& \multicolumn{5}{c|}{\textbf{Reasoning-Driven}}
& \multirow{2}{*}{\textbf{Overall}} \\
\cmidrule(lr){2-6} \cmidrule(lr){7-11}
& SE & Weather & MC & IP & WK
& SL & Poem & Life Reason & GU & Math
& \\
\midrule

GPT-Image-1~\citep{openai2024gptimage}
& 0.32 & 0.06 & 0.22 & 0.02 & 0.16
& 0.32 & 0.10 & 0.24 & 0.10 & 0.12
& 0.17 \\

GPT-Image-1.5~\citep{openai2025gptimage}
& 0.36 & 0.18 & 0.22 & 0.04 & 0.30
& 0.34 & 0.08 & 0.34 & 0.10 & 0.02
& 0.21 \\

FLUX 2 Pro~\citep{blackforestlabs2025flux2pro}
& 0.38 & 0.12 & 0.08 & 0.00 & 0.20
& 0.44 & 0.64 & 0.18 & 0.04 & 0.02
& 0.21 \\

FLUX 2 Max~\citep{blackforestlabs2025flux2max}
& 0.44 & 0.12 & 0.10 & 0.04 & 0.38
& 0.40 & 0.50 & 0.20 & 0.02 & 0.06
& 0.23 \\

Nano Banana~\citep{googledeepmind2025banana}
& 0.30 & 0.10 & 0.12 & 0.00 & 0.30
& 0.32 & 0.36 & 0.20 & 0.04 & 0.08
& 0.18 \\

Nano Banana Pro~\citep{googledeepmind2025bananapro}
& 0.50 & 0.36 & 0.40 & 0.16 & 0.56
& 0.62 & 0.68 & 0.30 & 0.16 & 0.46
& 0.41 \\

\midrule

SDXL~\citep{podell2023sdxl}
& 0.04 & 0.00 & 0.04 & 0.00 & 0.00
& 0.00 & 0.00 & -- & -- & --
& 0.01 \\

SD-3.5 M~\citep{stabilityai2024sd35medium}
& 0.02 & 0.00 & 0.00 & 0.00 & 0.02
& 0.00 & 0.00 & -- & -- & --
& 0.01 \\

SD-3.5 L~\citep{stabilityai2024sd35large}
& 0.04 & 0.00 & 0.02 & 0.00 & 0.02
& 0.00 & 0.06 & -- & -- & --
& 0.01 \\

FLUX 1 dev~\citep{labs2024flux}
& 0.04 & 0.00 & 0.00 & 0.00 & 0.02
& 0.02 & 0.04 & -- & -- & --
& 0.02 \\

FLUX 1 Kontext~\citep{labs2025fluxkontext}
& 0.02 & 0.00 & 0.00 & 0.00 & 0.02
& 0.00 & 0.00 & -- & -- & --
& 0.01 \\

FLUX 1 Krea~\citep{labs2024flux}
& 0.04 & 0.00 & 0.04 & 0.00 & 0.02
& 0.00 & 0.02 & -- & -- & --
& 0.02 \\

BAGEL~\citep{deng2025bagel}
& 0.02 & 0.00 & 0.00 & 0.00 & 0.00
& 0.02 & 0.02 & 0.02 & 0.00 & 0.08
& 0.02 \\

Echo-4o~\citep{ye2025echo4o}
& 0.04 & 0.00 & 0.00 & 0.00 & 0.00
& 0.02 & 0.06 & 0.02 & 0.02 & 0.02
& 0.02 \\

DraCo~\citep{jiang2025draco}
& 0.02 & 0.00 & 0.02 & 0.00 & 0.00
& 0.02 & 0.02 & 0.04 & 0.02 & 0.06
& 0.02 \\

Z-Image~\citep{cai2025zimage}
& 0.02 & 0.00 & 0.08 & 0.02 & 0.00
& 0.00 & 0.00 & -- & -- & --
& 0.02 \\

Qwen-Image~\citep{wu2025qwenimage}
& 0.08 & 0.00 & 0.04 & 0.00 & 0.00
& 0.04 & 0.00 & 0.04 & 0.00 & 0.00
& 0.02 \\

Mind-Brush~\citep{he2026mindbrush}
& 0.54 & 0.16 & 0.62 & 0.18 & 0.40
& 0.26 & 0.54 & 0.10 & 0.16 & 0.14
& 0.31 \\

\rowcolor{green!8}
\textbf{MemoGen (Ours)}
& \textbf{0.60} & 0.30 & 0.52 & \textbf{0.72} & \textbf{0.82}
& \textbf{0.74} & 0.60 & \textbf{0.42} & \textbf{0.20} & 0.30
& \textbf{0.52} \\

\bottomrule
\end{tabular}
}
\end{table*}

\section{Experiments}

\subsection{Benchmarks and Evaluation Protocols}

We evaluate MemoGen on two knowledge-intensive and reasoning-oriented image
generation benchmarks: WISE and Mind-Bench. WISE focuses on complex semantic
understanding and world-knowledge integration, covering categories such as
cultural knowledge, temporal reasoning, spatial reasoning, biology, physics, and
chemistry. We report category-level accuracy and overall accuracy following the
official evaluation protocol. Mind-Bench further evaluates image generation under
dynamic external knowledge and multi-step reasoning requirements, including both
knowledge-driven and reasoning-driven tasks. Following the benchmark protocol,
we report Checklist-based Strict Accuracy (CSA) for each subcategory and the
overall score. For both benchmarks, we report the results after the second
evolution round, where MemoGen has access to experience memory accumulated from
the previous round while keeping the underlying image generator fixed.
\begin{figure*}[t]
    \centering
    \includegraphics[width=0.9\linewidth]{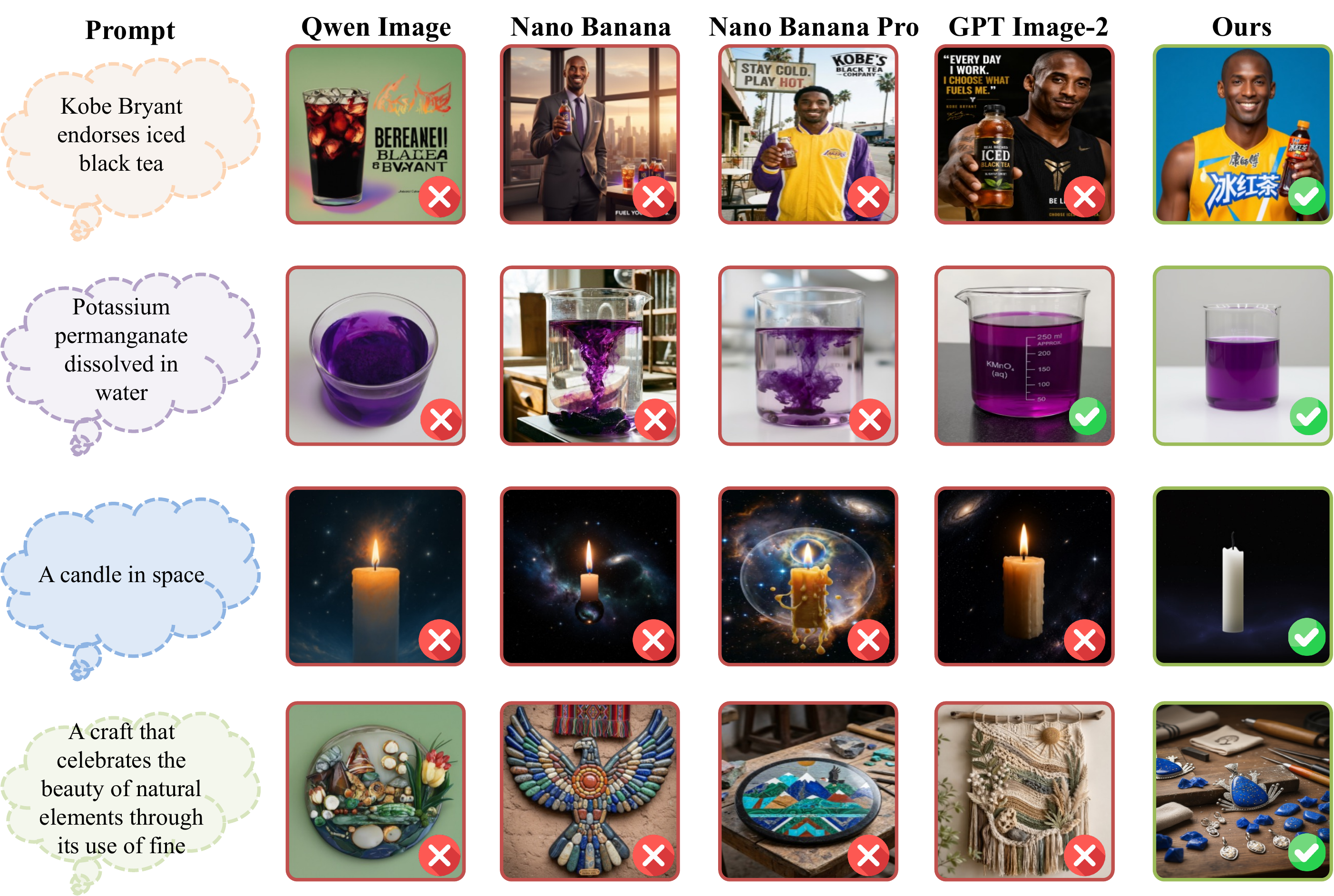}
    \caption{Qualitative comparison. The results in  illustrate that MemoGen ("Ours") effectively follows complex prompts and adheres to implicit constraints where other models struggle, such as correctly depicting an unlit candle in space or specific product endorsements.}
    \label{fig:vis-compare}
\end{figure*}
\subsection{Experimental Settings}

\paragraph{Baselines.}
We compare MemoGen with current mainstream proprietary image generation systems,
including GPT-Image-1~\citep{openai2025gptimage}, GPT-Image-1.5\citep{openai2025gptimage},
Nano Banana~\citep{googledeepmind2025banana}, Nano Banana
Pro~\citep{googledeepmind2025bananapro}, FLUX-2 Pro \citep{blackforestlabs2025flux2pro}, and FLUX-2 Max \citep{blackforestlabs2025flux2max}.
We also compare with state-of-the-art open-source text-to-image or unified
multimodal models, including FLUX.1-dev~\citep{labs2024flux}, FLUX 1 Kontext\citep{labs2025fluxkontext},
FLUX 1 Krea\citep{labs2025fluxkrea}, Stable Diffusion 3.5 Large~\citep{stabilityai2024sd35large},
Stable Diffusion 3.5 Medium~\citep{stabilityai2024sd35medium},
SDXL~\citep{podell2023sdxl}, BAGEL~\citep{deng2025bagel},
Echo-4o\citep{ye2025echo4o} , DraCo\citep{jiang2025draco}, Z-Image\citep{cai2025zimage}, GenAgent~\citep{jiang2026genagent}, and
Qwen-Image~\citep{wu2025qwenimage}. In addition, we include two closely related
agentic generation systems, Mind-Brush\citep{he2026mindbrush}  and Gen-Searcher\citep{feng2026gensearcher} , to compare MemoGen
against retrieval-augmented and search-augmented image generation frameworks.
All baselines are evaluated using their official default settings or reported
numbers when available.

\paragraph{Implementation details.}
Unless otherwise specified, MemoGen uses Qwen-Image-Edit-2509\citep{wu2025qwenimage} as the
reference-guided generation backend and Qwen-Image as the text-only generation
backend. GPT-5.5\citep{openai2025gpt55} is used as the VLM controller for intent analysis,
tool planning, visual inspection, and memory-guided feedback. For external
retrieval, we use Serper\citep{serper2025} as the search interface. The local visual RAG module is
built on the WIT image-text corpus~\citep{srinivasan2021wit}, which provides large-scale image-caption
pairs for efficient local reference retrieval. The agent is equipped with local
visual RAG, dynamic web search, reference retrieval, experience memory
retrieval, image generation, visual inspection, and memory writeback tools.
For WISE and Mind-Bench, we report the second-round evolution results, where the
agent retrieves experience records accumulated from the first round and performs
memory-guided generation without updating the parameters of the image generator
or controller. All experiments are conducted on 8 NVIDIA A100 80GB GPUs.
Importantly, all evolution is performed at test time through memory retrieval
and memory writeback; neither the image generator nor the VLM controller is
fine-tuned during the evolution process.

\subsection{Comparison with State of the Art}

We evaluate MemoGen on Mind-Bench and WISE, two complementary benchmarks for
knowledge-intensive and reasoning-oriented image generation. Mind-Bench stresses
dynamic external knowledge and multi-step reasoning, while WISE focuses on
translating world knowledge into faithful visual content across cultural,
temporal, spatial, biological, physical, and chemical scenarios. For both
benchmarks, we report second-round results with the image generator kept fixed;
thus, any improvement comes from retrieving and reusing experience accumulated
from previous generation episodes.

Table~\ref{tab:mindbench_results} shows that MemoGen achieves the best overall
performance on Mind-Bench, reaching an overall CSA of \textbf{0.52}. This
outperforms the strongest proprietary baseline, Nano Banana Pro
(\textbf{0.41}), and substantially exceeds Mind-Brush (\textbf{0.31}).
The comparison with Mind-Brush is especially informative: both methods use
external knowledge and agentic reasoning, but MemoGen additionally reuses past
successes and failures as explicit experience. The \textbf{0.21} absolute gain
therefore suggests that experience memory provides benefits beyond single-round
retrieval and prompt refinement. At the category level, MemoGen performs strongly
on both Knowledge-Driven tasks, such as SE, IP, and WK
(\textbf{0.60}, \textbf{0.72}, and \textbf{0.82}), and Reasoning-Driven tasks,
such as SL, Life Reasoning, and GU (\textbf{0.74}, \textbf{0.42}, and
\textbf{0.20}). This indicates that memory records do not merely recall facts,
but help convert past generation outcomes into reusable visual constraints,
repair hints, and failure-avoidance signals.

Table~\ref{tab:wise_results} further confirms this trend on WISE. MemoGen
achieves the best overall score of \textbf{0.91}, outperforming strong
proprietary systems such as GPT-Image-1, Nano Banana, and Nano Banana Pro, as
well as agentic baselines including Gen-Searcher and Mind-Brush. Compared with
Mind-Brush (\textbf{0.78}) and Gen-Searcher (\textbf{0.77}), MemoGen improves
the overall score by \textbf{0.13} and \textbf{0.14} absolute points,
respectively. The category-level results show that MemoGen reaches
\textbf{0.96} on Cultural, \textbf{0.90} on Time, and \textbf{0.85} on
Chemistry. These categories are prone to recurring visual errors: a model may
infer the correct answer textually, yet still render the wrong time of day,
physical state, or intermediate concept. By retrieving relation-level experience
from previous episodes, MemoGen turns such recurring failures into concrete
generation constraints.

Overall, the two benchmarks support the same conclusion from different angles:
search-augmented systems improve the current prompt by acquiring external
knowledge, while MemoGen treats each generation episode as a source of future
experience. Successful records provide candidate strategies, and failed or
rejected records become warnings and repair constraints. As a result, the system
improves through training-free continual learning at the agent layer, rather
than through fine-tuning the image generator.

Figure~\ref{fig:vis-compare} presents qualitative comparisons between MemoGen
and representative baseline models across personality, chemistry, physics, and
cultural/geographical reasoning scenarios. In the first example, ``Kobe Bryant
endorses iced black tea,'' baseline models either fail to correctly ground the
celebrity identity or misunderstand the visual concept of iced black tea as a
Chinese beverage. In contrast, MemoGen successfully preserves both key elements
and further infers the underlying Internet-meme association, rendering the
result in a recognizable meme-like visual form. In the chemistry example, only
MemoGen correctly captures the color and physical state of potassium
permanganate dissolved in water, producing a homogeneous purple solution rather
than undissolved crystals or an incorrect precipitate-like state. In the physics
example, ``A candle in space,'' most baselines follow the strong visual prior
that candles are naturally associated with flames, and therefore generate
physically implausible images of a burning candle in space. By reasoning that
combustion cannot be sustained in vacuum and by incorporating prior failure
experience, MemoGen applies explicit negative constraints to suppress the
candle--flame association, ultimately generating a physically consistent unlit
candle with an inert wick. These examples show that MemoGen improves not only
factual grounding but also the visual realization of implicit reasoning,
especially when the task requires combining external knowledge, commonsense
constraints, and failure-avoidance experience.

\begin{figure}
    \centering
    \includegraphics[width=\linewidth]{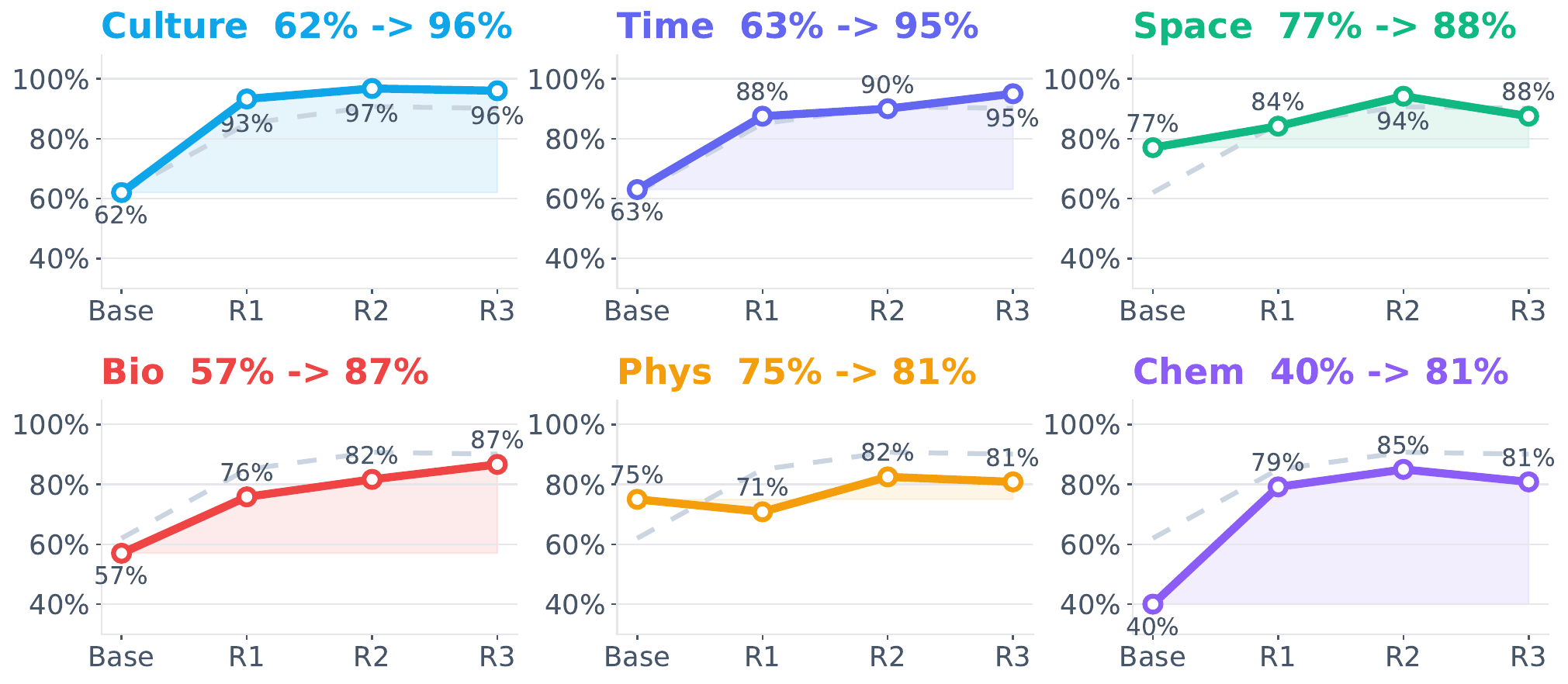}
    \caption{Category-wise WISE pass rates from the Qwen Image baseline to three evolving rounds. The dashed gray line indicates the overall score trend.}
    \label{fig:placeholder}
\end{figure}

\paragraph{Evolution analysis.}
We further analyze the effect of evolution rounds in Figure~\ref{fig:placeholder}.
The improvement from the base generator to Round~1 mainly reflects the benefit
of our agentic generation architecture: even without task-specific experience
memory, the controller can actively decompose the prompt, retrieve external
knowledge and visual references, construct explicit generation constraints, and
review the generated image. This already leads to substantial gains over the
base model. The further improvement from Round~1 to Round~2 demonstrates the
value of experience-driven evolution. By retrieving relation-level memories
accumulated in the first round, the agent can better exploit the strengths of
the underlying generator while avoiding previously observed failure modes, such
as missing key visual cues, incorrect temporal states, ambiguous object
relations, or physically implausible renderings. The performance of Round~3 is
largely comparable to Round~2, with small category-level fluctuations. We
attribute this saturation to two factors: some remaining cases are limited by
the intrinsic capability of the fixed image generator, while others are affected
by noise or misjudgments from the automatic evaluator. Overall, the evolution
curve shows that self-evolution is beneficial: the largest gain comes from
introducing agentic control, and experience memory further stabilizes and
improves generation in later rounds without updating model parameters.
\section{Conclusion}

We introduced MemoGen, a training-free continual learning framework for
text-to-image generation agents. The central insight is that future image
generation systems need not improve only by scaling or fine-tuning the underlying
generator; they can also improve by accumulating and reusing experience at the
agent layer. By converting each generation episode into explicit memory,
including task understanding, retrieved evidence, visual feedback, successful
strategies, and failure constraints, MemoGen turns image generation from an
isolated prompt-to-image inference process into an experience-evolving cycle.
Experiments on knowledge-intensive and reasoning-oriented benchmarks show that
past generation experience can serve as an effective test-time learning signal,
helping the system construct better generation conditions, avoid recurring
errors, and improve reliability without parameter updates. We believe this
experience-evolving paradigm complements retrieval-augmented generation and
points toward future visual generation systems that continuously improve through
interaction, feedback, and accumulated experience.

\clearpage
\printbibliography
\end{document}